\documentclass[10pt,twocolumn,letterpaper]{article}

\usepackage{iccv}
\usepackage{times}
\usepackage{epsfig}
\usepackage{graphicx}
\usepackage{amsmath}
\usepackage{amssymb}
\usepackage{subcaption,siunitx,booktabs}
\usepackage[accsupp]{axessibility}

\usepackage{pifont}
\newcommand{\cmark}{\ding{51}}%
\newcommand{\xmark}{\ding{55}}%

\newcommand\blfootnote[1]{%
  \begingroup
  \renewcommand\thefootnote{}\footnote{#1}%
  \addtocounter{footnote}{-1}%
  \endgroup
}

\usepackage[pagebackref=true,breaklinks=true,letterpaper=true,colorlinks,bookmarks=false]{hyperref}

\iccvfinalcopy 

\def\iccvPaperID{32} 

\ificcvfinal\fi

\begin{document}

\title{\textit{Lizard}: A Large-Scale Dataset for Colonic Nuclear Instance Segmentation and Classification}


\author{Simon Graham$^1$, Mostafa Jahanifar$^1$, Ayesha Azam$^2$, Mohammed Nimir$^2$, Yee-Wah Tsang$^2$, \\ Katherine Dodd$^2$, Emily Hero$^{2,3}$, Harvir Sahota$^2$, Atisha Tank$^2$, Ksenija Benes$^4$, Noorul Wahab$^1$, \\ Fayyaz Minhas$^1$, Shan E Ahmed Raza$^1$, Hesham El Daly$^2$, Kishore Gopalakrishnan$^2$, \\ David Snead$^2$, Nasir Rajpoot$^1$ \\ \\
$^1$Department of Computer Science, University of Warwick, UK \\
$^2$Department of Pathology, University Hospitals Coventry and Warwickshire NHS Trust, UK \\
$^3$Department of Pathology, University Hospitals of Leicester NHS Trust, UK \\
$^4$Department of Pathology, The Royal Wolverhampton NHS Trust, UK
\\
{\tt\small simon.graham@warwick.ac.uk}
}

\maketitle
\thispagestyle{empty}

\begin{abstract}
The development of deep segmentation models for computational pathology (CPath) can help foster the investigation of interpretable morphological biomarkers. Yet, there is a major bottleneck in the success of such approaches because supervised deep learning models require an abundance of accurately labelled data. This issue is exacerbated in the field of CPath because the generation of detailed annotations usually demands the input of a pathologist to be able to distinguish between different tissue constructs and nuclei. Manually labelling nuclei may not be a feasible approach for collecting large-scale annotated datasets, especially when a single image region can contain thousands of different cells. However, solely relying on automatic generation of annotations will limit the accuracy and reliability of ground truth. Therefore, to help overcome the above challenges, we propose a multi-stage annotation pipeline to enable the collection of large-scale datasets for histology image analysis, with pathologist-in-the-loop refinement steps. Using this pipeline, we generate the largest known nuclear instance segmentation and classification dataset, containing nearly half a million labelled nuclei in H\&E stained colon tissue. We have released the dataset and encourage the research community to utilise it to drive forward the development of downstream cell-based models in CPath.

\end{abstract}

\section{Introduction}

Deep learning models have revolutionised the field of computational pathology (CPath), partly due to their ability in leveraging the huge amount of image data contained in multi-gigapixel whole-slide images (WSIs). In particular, convolutional neural networks (CNNs) have shown great promise and have been successfully applied to various tasks including cancer screening \cite{bejnordi2017diagnostic, campanella2019clinical}, cancer grading \cite{bulten2020automated, shaban2020context} and cancer type prediction \cite{graham2018classification, lu2020deep}. Yet, utilising CNNs in an end-to-end manner for slide-level prediction can lead to poor explainability, due to a high-level of model complexity with limited feature interpretability \cite{tosun2020explainable, holzinger2017towards, jaume2020quantifying}. Explainable AI (or XAI) in CPath may be preferable because it can ensure algorithmic fairness, identify potential bias in the training data, and ensure that the algorithms perform as expected \cite{gilpin2018explaining}. Furthermore, XAI using human-interpretable features \cite{diao2021human} can help uncover novel biomarkers for complex tasks, such as predicting genetic alterations using tissue morphology alone \cite{kather2020pan}. 
\blfootnote{© 2021 IEEE.  Personal use of this material is permitted.  Permission from IEEE must be obtained for all other uses, in any current or future media, including reprinting/republishing this material for advertising or promotional purposes, creating new collective works, for resale or redistribution to servers or lists, or reuse of any copyrighted component of this work in other works.}

 	\begin{figure*}[t]
		\centering
        \includegraphics[width=1.0\textwidth]{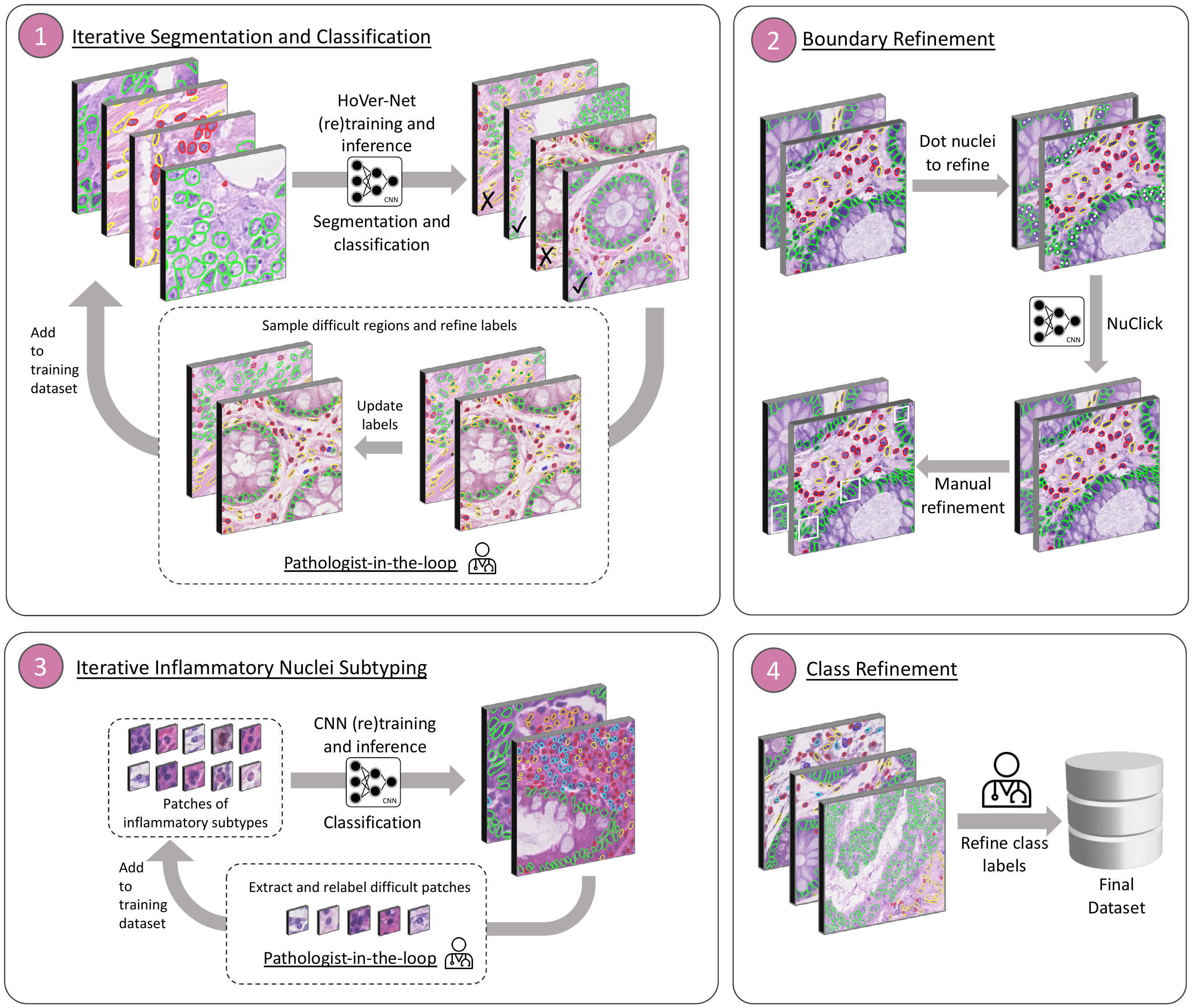}
		\caption{Overall pipeline to collect annotations for Lizard.} 
		\label{fig:pipeline}
	\end{figure*}

In order to extract meaningful human-interpretable features from the tissue, accurate localisation of clinically-relevant structures is an important initial step. For example, features indicative of nuclear morphology first require each nucleus to be segmented and can then be directly used in downstream tasks, such as predicting the cancer grade \cite{alsubaie2018bottom}. Recently, it has been shown that CNNs achieve state-of-the-art performance in CPath for segmentation of structures such as glands \cite{graham2019mild,chen2017dcan,graham2019rota}, nuclei \cite{naylor2018segmentation,graham2019hover, zhao2020triple}, nerves and blood vessels \cite{fraz2020fabnet}. However, these deep segmentation models are very \textit{data hungry} and consequently need a large amount of pixel-level annotations to perform well \cite{tizhoosh2018artificial}. This data collection is a major bottleneck in the development of deep learning algorithms for segmentation tasks in CPath because it requires the input of domain experts and accurately delineating object boundaries is a very time consuming procedure. There has been recent progress in the development of large-scale nuclear instance segmentation datasets in computational pathology \cite{gamper2020pannuke, amgad2021nucls}, but there exists a risk that as we increase the amount of data, there becomes a trade-off with the accuracy of the collected annotations. Furthermore, quantitative multi-pathologist evaluation becomes difficult when there exists hundreds of thousands of different annotated cells.

Therefore, to tackle the challenges described above, in this paper we introduce:

\begin{itemize}
    \item A multi-stage annotation pipeline to enable the collection of accurate large-scale instance segmentation datasets.
    \item The largest instance segmentation and classification dataset in computational pathology containing nearly half a million labelled nuclei.
    \item A novel method to gain accurate quantitative concordance statistics for a representative sample of our dataset.
\end{itemize}

Our developed dataset, which we call \textbf{Lizard}\footnote{Download dataset at: \url{warwick.ac.uk/lizard-dataset}}, consists of histology image regions of colon tissue from six different dataset sources at $\pmb{20\times}$ objective magnification, with full segmentation annotation for different types of nuclei. In particular we provide the nuclear class label for epithelial cells, connective tissue cells, lymphocytes, plasma cells, neutrophils and eosinophils and therefore models trained on this data may be used to help effectively profile the colonic tumour micro-environment. We choose to focus on nuclei from colon tissue to ensure that our dataset contains images from a wide variety of different normal, inflammatory, dysplastic and cancerous conditions in the colon - therefore increasing the likelihood of generalisation to unseen examples. Lizard is generated using a multi-stage approach, where focus is directed upon the refinement of below satisfactory automatic and semi-automatic predictions to enable the collection of large-scale annotated datasets. 
    
 \begin{table}[t!]
\centering
\begin{tabular}{|l|c|c|c|}
\hline
 \textbf{Dataset}  & \textbf{Nuclei} & \textbf{Labels} & \textbf{Organs}    \\ \hline
CPM-17 \cite{vu2019methods} & 7,570 &  \xmark & Multiple            \\ \hline  
TNBC \cite{naylor2018segmentation} & 4,056 & \xmark & Breast          \\ \hline  
MoNuSeg \cite{kumar2019multi} & 21,623 & \xmark & Multiple           \\ \hline  
MoNuSAC \cite{verma2021monusac2020} & 46,909 & \cmark & Multiple            \\ \hline  
CoNSeP \cite{graham2019hover} & 24,319 & \cmark & Colon           \\ \hline  
Colon PanNuke \cite{gamper2020pannuke} & 35,711 & \cmark & Colon         \\ \hline  
PanNuke \cite{gamper2020pannuke} & 205,343 & \cmark & Multiple         \\ \hline  
NuCLS \cite{amgad2021nucls} & 222,396 & \cmark & Breast         \\ \hline  
Lizard & 495,179 & \cmark & Colon            \\ \hline  
\end{tabular}
\caption{Existing nuclei segmentation datasets.}
\label{table:nuclei-datasets}
\end{table}


\section{Related work}
\subsection{Nuclear segmentation data}
Recently, a substantial effort has been spent creating accurately labelled datasets for the development of nuclear segmentation machine learning algorithms. Initial work focused on labelling all nuclei as a single category. For example, the TNBC dataset \cite{naylor2018segmentation} consists of triple negative breast cancer image regions, where nuclear boundaries are fully labelled. Also, the CPM-17 and MoNuSeg datasets \cite{vu2019methods,kumar2019multi} label all nuclear boundaries of image regions extracted from multiple tissues within the TCGA database. However, the above datasets were annotated manually, which limits the number of nuclei that can be labelled at scale.

\subsection{Nuclear segmentation and classification data}
The above datasets have been pivotal in enabling the development of nuclear segmentation methods in CPath, but they do not consider the nuclei as separate categories. Accurate prediction of the type of each nucleus is beneficial because it can subsequently enable profiling of the tumour micro-environment (TME). 

CoNSeP \cite{graham2019hover} was one of the first open source datasets to provide the segmentation boundaries of nuclei along with their associated class labels, enabling the development of simultaneous nuclear segmentation and classification approaches in CPath, such as HoVer-Net \cite{graham2019hover}. Similarly, MoNuSAC \cite{verma2021monusac2020} provides the segmentation mask and class labels for nuclei extracted from TCGA and also sub-categorises inflammatory cells as either lymphocyte, neutrophil, eosinophil or macrophage. Therefore models trained on this dataset may be strong indicators on the status of the TME. CoNSeP and MoNuSAC annotations were generated manually and therefore similar protocols will not scale for large-scale nuclear segmentation and classification datasets. 

To help overcome the challenge of annotating a large number of different nuclei, PanNuke \cite{gamper2020pannuke} employed a semi-automatic approach with several quality control steps. Despite providing a remarkable number of labelled nuclei, PanNuke relies on semi-automatic generation of nuclear boundaries and therefore results may be unsatisfactory for a number of difficult cases. Furthermore, due to the large number of different tissue types considered, each tissue dependent subset will have significantly fewer nuclei, where nuclear morphology may be different.  

Amgad \textit{at al.} \cite{amgad2021nucls} developed an approach to help speed up the annotation collection process of nuclei within tissue regions containing breast cancer. The annotation protocol leverages the knowledge collected via crowd sourcing to provide accurately labelled nuclear classes. However, automatic generation of all nuclear boundaries was relied upon, which may lead to inaccurate annotation for challenging nuclei.

A full breakdown of the numbers of nuclei within each of the mentioned datasets is provided in Table \ref{table:nuclei-datasets}. Here, we observe that Lizard is the largest nuclear instance segmentation dataset, where the label for each individual nucleus is also provided. 

  	\begin{figure*}[t]
		\centering
        \includegraphics[width=1.0\textwidth]{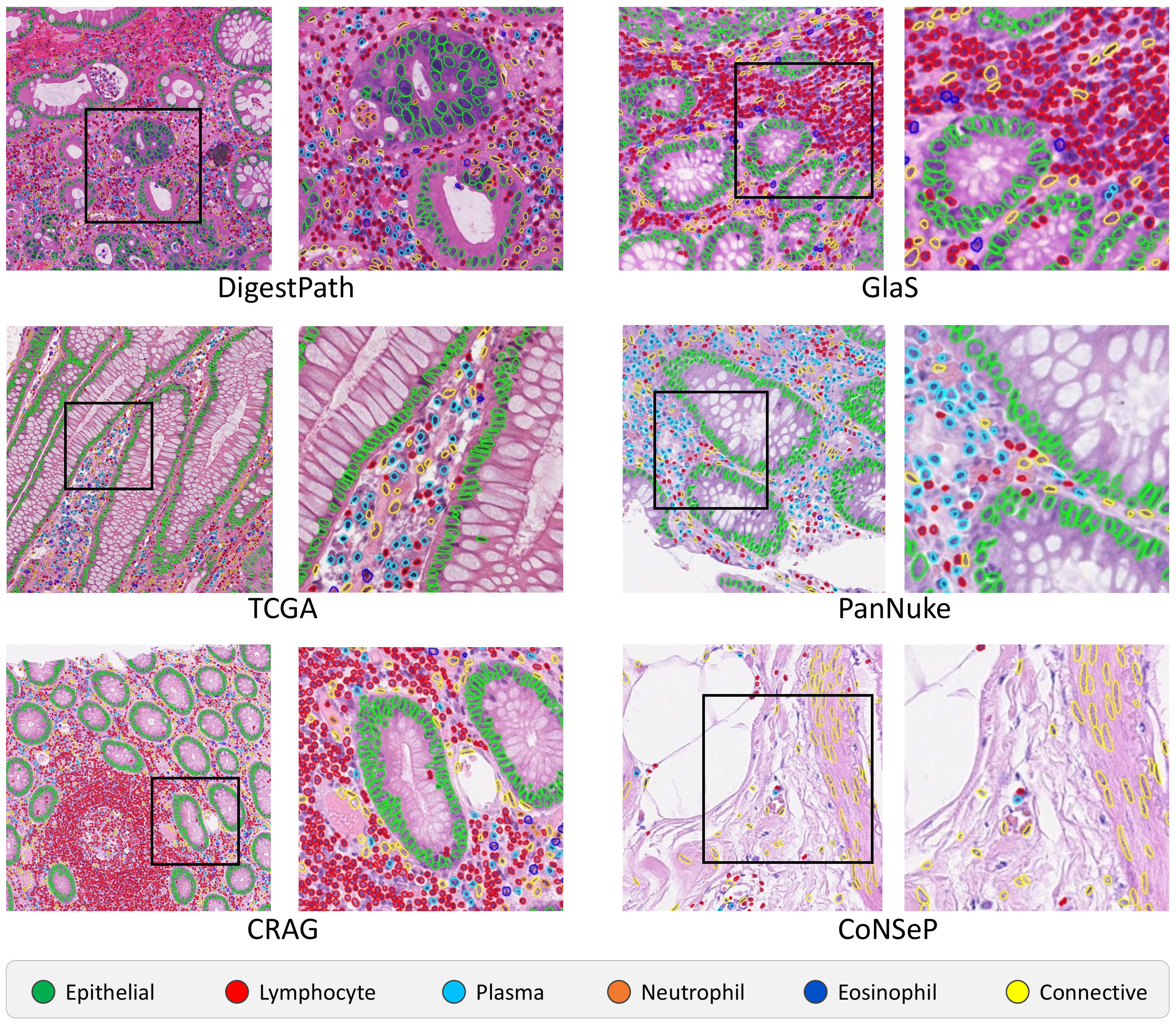}
		\caption{Example image regions sampled from each of the six data sources.} 
		\label{fig:dataset}
	\end{figure*}
	
\section{The dataset generation pipeline}
Our pipeline for collecting accurate nuclei annotations at scale consists of four steps. First we optimise an automated segmentation and classification method to annotate unlabelled nuclei as belonging to either epithelial, inflammatory or connective tissue cells. Then, we apply a sequential boundary refinement strategy, where attention is only given to nuclei with unsatisfactory results. Next we apply an automated method to sub-type inflammatory nuclei as either lymphocyte, plasma, neutrophil or eosinophil and finally the class labels of all nuclei are verified. During our entire pipeline, there is significant pathologist interaction to ensure our final dataset is reliable. An overview of the proposed pipeline can be seen in Figure \ref{fig:pipeline}.

In the sections below, we will initially describe the data sources and then provide further detail into each of the aforementioned steps.

\subsection{Data sources}
To generate our final dataset, we considered data from 6 different sources: GlaS \cite{sirinukunwattana2017gland}, CRAG \cite{graham2019mild}, CoNSeP \cite{graham2019hover}, DigestPath, PanNuke \cite{gamper2020multi} and TCGA \cite{grossman2016toward}. GlaS, CRAG and CoNSeP contain image regions extracted from whole-slide images (WSIs) from University Hospitals Coventry and Warwickshire (UHCW), whereas TCGA contains WSIs from multiple centres in the USA. PanNuke contains image regions originating from both UHCW and TCGA, while DigestPath contains image regions extracted from biopsy samples from 4 different Chinese hospitals. To develop Lizard, we sampled colon image regions from the original data sources at 20$\times$ objective magnification ($\sim$0.5$\mu$m/pixel). This resolution was used because it was the maximum available magnification for the GlaS, CRAG and DigestPath datasets. For both CoNSeP and PanNuke, we only utilised the images and not the associated annotations from the original datasets to ensure that the same label generation pipeline was used on all input data. In total, we extracted 291 image regions with an average size of 1,016$\times$917 pixels from a large number of different patients. Sample image regions from each of the used data sources can be seen in Figure \ref{fig:dataset}.

\subsection{Iterative segmentation and classification}
\label{section:iterative-seg}
As a first step, we leveraged publicly available data and trained an automated nuclear segmentation and classification model to label nuclei in an unannotated target dataset. In particular, we trained HoVer-Net \cite{graham2019hover}, which is a state-of-the-art algorithm designed specifically for this purpose, on the colon subset of the PanNuke dataset \cite{gamper2020pannuke} and segmented nuclei as either epithelial, inflammatory or belonging to connective tissue. Here, connective tissue cells comprise of a broader category containing nuclei from fibroblasts, muscle and endothelial cells. We decided to group neoplastic and non-neoplastic epithelial nuclei into a single category to help ease our downstream class refinement pipeline. 

Our aim was to ensure that our algorithm performed as well as possible on our target dataset that we wish to label. Therefore, with the assistance of two pathologists, we extracted small image regions of size 512$\times$512 pixels from our target dataset where the HoVer-Net results were poor and manually refined the segmentation boundaries and class predictions. These image regions along with their refined annotations were then added to the initial training dataset and the model was retrained. We repeated this procedure for several steps until we were satisfied with the results.

\subsection{Boundary refinement}
As the iterative relabelling and retraining of our automatic model was done according to the predictions made by HoVer-Net, there will still inevitably be nuclei with a poor segmentation result, especially for challenging cases with indistinct boundaries. In the second step of our pipeline, we applied a sequential approach for boundary refinement of the labelled nuclei. Our philosophy here is that we should only need to manually trace the object boundaries when automatic / semi-automatic results are not good enough. Using this approach, we significantly reduce the number of nuclei requiring manual annotation, enabling the collection of accurate annotations at scale.

\textbf{Semi-automatic refinement:} In the first step of boundary refinement, we provided a single point annotation centred at each nucleus where we observed a poor quality segmentation result. Next, we generated the refined nuclear boundaries using NuClick \cite{koohbanani2020nuclick}, which uses each point annotation as a guiding signal to achieve a superior segmentation compared to fully automated methods. 

\textbf{Manual refinement:} Despite NuClick easing the refinement process, there will still be imperfections in the results that we may wish to refine. As a final boundary refinement step, we manually refined any NuClick results that we believed should be further improved. Hence, using this strategy we significantly reduced the number of nuclei that required their entire nuclear boundary to be manually delineated. To speed up the collection of annotations, the manual refinement step can be skipped, but may come at the cost of reduced annotation accuracy. 

\subsection{Iterative inflammatory nuclei sub-typing}
\label{section:inf-subtype}
Sub-categorisation of inflammatory nuclei may be used to better profile the tissue micro-environment and enables further exploration of the link between different types of inflammatory cells and patient outcome. In recent years, tumour infiltrating lymphocytes (TILs) have been shown to be associated to prognosis \cite{fridman2012immune}, yet further downstream analysis of the complex interaction between additional inflammatory cells may provide us with a stronger indicator. Therefore, we developed an automated model to sub-type the inflammatory cells as either lymphocyte, plasma, neutrophil or eosinophil.

To train our inflammatory cell sub-typing algorithm, we leveraged 71,112 nuclei from an internal colon dataset at the University Hospitals Coventry and Warwickshire (UHCW), where each nuclear category was given by a team of pathologists. More specifically, this dataset comprised of 54,850 lymphocytes, 9,821 plasma cells, 3,666 neutrophils and 2,775 eosinophils. We then extended our inflammatory sub-typing dataset by considering 15,654 lymphocytes and 631 neutrophils from the MoNuSAC \cite{verma2021monusac2020} dataset. Then, we extracted an image patch of size 32 $\times$ 32 pixels at 20$\times$ centred at each labelled nucleus to form our initial dataset of inflammatory nuclei patches. 

 Using this patch-level dataset, we trained a simple CNN consisting of 6 convolution layers, 2 max pooling operations and a fully connected layer to predict the category of each inflammatory nucleus. We then processed patches centred at each inflammatory nucleus within our target dataset to predict the corresponding subtype. Similar to the procedure explained in Section \ref{section:iterative-seg}, a random sample of miss-classified nuclei was then extracted from the target dataset, as directed by pathologists, which was added to the original training dataset before retraining. This strategy was repeated until the results of the inflammatory sub-typing model performed at a satisfactory level. 

\subsection{Class refinement}
As a final step, we refined the class labels of all nuclei with the assistance of two pathologists. For this, all of the images from our target dataset and their corresponding annotations were ported into a web-based annotation application and class labels were updated via a series of virtual refinement sessions. A single click was provided within any nucleus that required its class to be updated with a specified new class.

 	\begin{figure}[t]
		\centering
        \includegraphics[width=1.0\columnwidth]{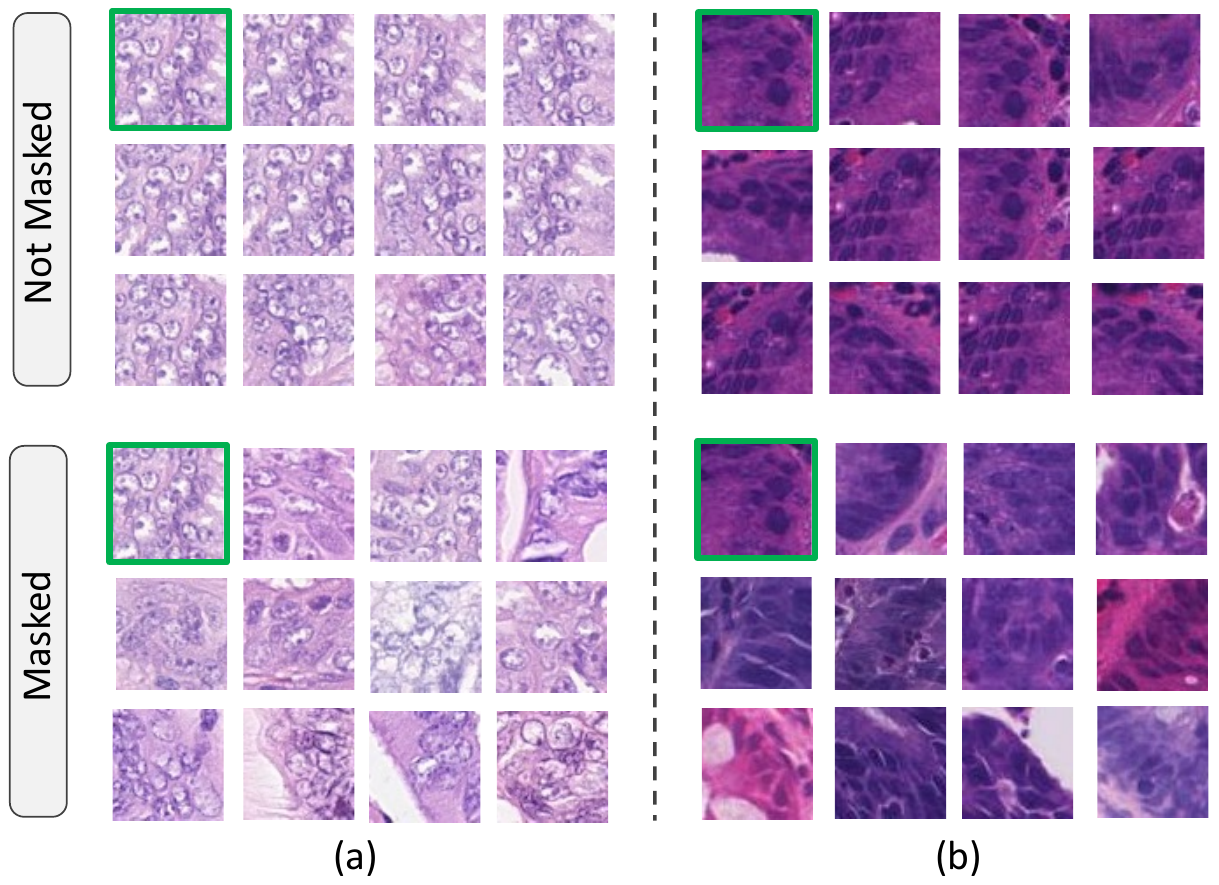}
		\caption{Nearest neighbours to the features of two query images using contrastive learning. The bottom row performs a masking operation to the input image, which ensures features focus on the central nucleus.} 
		\label{fig:nbors}
	\end{figure}
	
\section{Dataset statistics and baseline results}
\subsection{Contrastive learning for representative samples}
\label{section-contrastive}
Inspired by batch testing, we selected a representative sample from our developed dataset to assess the quality of the annotations. This was done by ensuring that the level of disagreement between a single pathologist and our final dataset labels was no greater than the intra-observer disagreement between multiple pathologists. 

To obtain some representative sample nuclei, we first trained a ResNet-34 network with contrastive learning \cite{chen2020simple} with image patches of size 64$\times$64 pixels extracted at the centre of each nucleus. We used a slightly larger patch size than that described in Section \ref{section:inf-subtype} to account for the fact that epithelial cells are typically larger than inflammatory cells. Contrastive learning is a recent state-of-the-art strategy that enables representative image features to be learned in an unsupervised setting by encouraging two transformed copies of an image to have similar feature embeddings. To ensure that the features learned are indicative of nuclear morphology and not the surrounding information, we applied a masking operation and only considered pixels within the immediate neighbourhood of each central nucleus. In particular, we dilated the binary nuclear mask and multiplied this with the original image. In Figure \ref{fig:nbors}, we show the nearest neighbours to the features of a query image (images with green borders) learned with and without the masking operation. It can be seen from this figure that masking ensures that features close together in the feature space have a similar nuclear appearance. In particular, the images in (a) all contain a central epithelial cell with a visible nucleolus, whereas in (b) each patch contains a central epithelial cell with a relatively homogeneous stain appearance. If masking is not used, then features relate more to the \textit{overall} appearance of the image.

After obtaining the features via contrastive learning, we apply hierarchical agglomerative clustering to each class independently to obtain 15 clusters per class giving us a total of 90 different clusters. A similar clustering method was used by Feng \textit{et al.} \cite{feng2021nuc2vec}, which implies that our clustering technique may be subsequently used for further fine-grained sub-typing of the nuclei in our dataset. Then, we select a random sample of 20 nuclei per cluster to give us an overall sample size of 1,800. We display 2 random clusters per category in Figure \ref{fig:clusters}.

\subsection{Pathologist assessment and concordance}
The representative samples generated as described in the above section were then provided to three different pathologists in a random order as a 96$\times$96 pixel image region with the nucleus at the centre. The pathologists reported the category of each nucleus and subsequently concordance statistics between the pathologists and Lizard were computed. Specifically, we report the Cohen's kappa ($\kappa$) between all possible pairs and provide the margin of error corresponding to the 95\% confidence interval \cite{fleiss1969large}. These results are displayed in Table \ref{table:concordance}, where pathologist A is a consultant pathologist and pathologists B and C are trainee pathologists. Our representative sample guarantees that there exist nuclei with a wide variety of morphological appearances from different classes and therefore the reported concordance statistics provide us with a good indication of the accuracy of the annotated data.

	\begin{figure}[b]
		\centering
        \includegraphics[width=1.0\columnwidth]{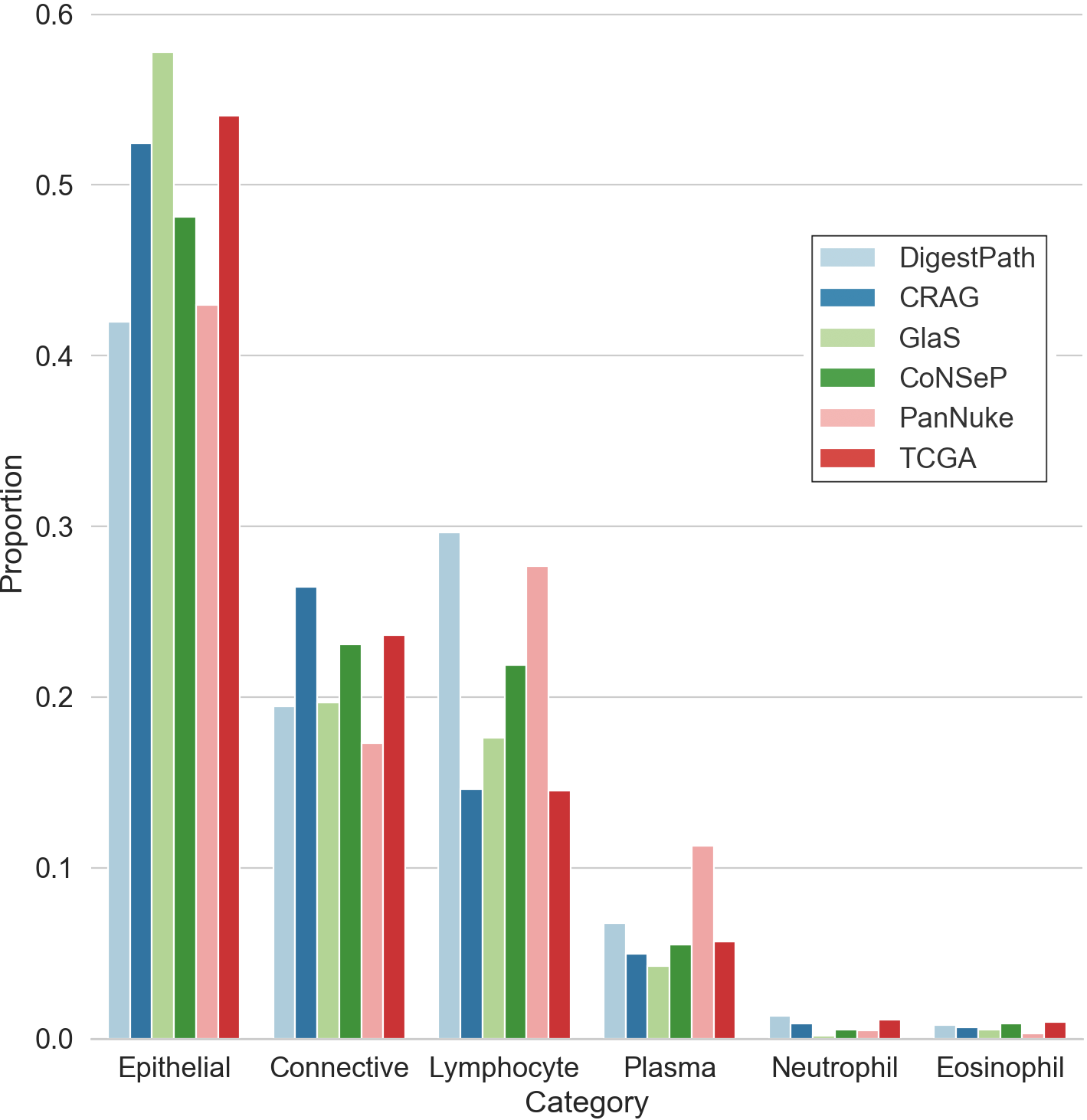}
		\caption{Distribution of categories in the overall dataset.} 
		\label{fig:dist}
	\end{figure}
	

\begin{table}[h]
\centering
\begin{tabular}{|c|c|}
\hline \textbf{Setting}   &  $\pmb{\kappa}$\\ 
\hline Pathologist A vs Pathologist B  & 0.6207 $\pm$ 0.0294 \\ 
\hline Pathologist A vs Pathologist C  & 0.6454 $\pm$ 0.0279   \\ 
\hline Pathologist B vs Pathologist C & 0.6086 $\pm$ 0.0286  \\ 
\hline Pathologist A vs Lizard  & 0.6420 $\pm$ 0.0278  \\ 
\hline Pathologist B vs Lizard  & 0.6373 $\pm$ 0.0282  \\ 
\hline Pathologist C vs Lizard & 0.6733 $\pm$ 0.0270   \\ 
\hline
\end{tabular}
\caption{Concordance between different pathologists and our generated dataset on a representative sample.}
\label{table:concordance}
\end{table}

From Table \ref{table:concordance}, we can see from the kappa scores that there is substantial agreement \cite{cohen1960coefficient} in all settings. We also observe that the concordance between each pathologist and our dataset is not significantly different to the pathologist concordance between each other. Therefore, we can be confident that the annotations in Lizard are accurate and the level of observed disagreement is well justified.

\begin{table*}[h]
\centering
\begin{tabular}{|l|c|c|c|c|c|c|c|}
\hline  &  \textbf{DigestPath} &  \textbf{CRAG} & \textbf{GlaS} & \textbf{PanNuke} & \textbf{CoNSeP}  & \textbf{TCGA}  &  \underline{\textbf{Total}} \\ 
\hline \textbf{Epithelial}  & 70,789 & 99,124 & 31,986 & 5,575 & 2,898 & 34,191 & 244,563 \\ 
\hline \textbf{Lymphocyte}  & 49,932 & 27,634 & 9,763 & 3,592 & 1,317 & 9,175 & 101,413  \\ 
\hline \textbf{Plasma}  & 11,352 & 9,363 & 2,349 & 1,465 & 332 & 3,605 & 28,466 \\ 
\hline \textbf{Neutrophil}  & 2,262 & 1,673 & 90 & 61 & 30 & 708 & 4,824 \\ 
\hline \textbf{Eosinophil}  & 1,349 & 1,255 & 286 & 37 & 52 & 625 & 3,604 \\ 
\hline \textbf{Connective} & 32,826 & 49,994 & 10,890 & 2,248 & 1,389 & 14,962 & 112,309 \\ 
\hline \underline{\textbf{Total}} & 168,510 & 189,043 & 55,364 & 12,978 & 6,018 & 63,266 & \textbf{495,179} \\ 
\hline
\end{tabular}
\caption{Summary of the annotations present in the Lizard dataset.}
\label{table:counts}
\end{table*}

\subsection{First order statistics}
In Table \ref{table:counts} we display the full breakdown of the number of labelled nuclei from each dataset source. We can see from this table that the majority of labelled nuclei come from the DigestPath and CRAG datasets, whereas the fewest labelled nuclei come from the CoNSeP dataset. It is important to note that we do not use \textit{all} of the the original images along with their corresponding annotations from PanNuke and CoNSeP and therefore numbers do not necessarily align with the original datasets. In addition, we highlight the distribution of nuclei counts in Figure \ref{fig:dist} for each data source used in our generated dataset. Here, we can see that a similar distribution of nuclei types are present across the different data sources, which is to be expected because it reflects how often these nuclei appear in the tissue. For example, there are significantly fewer eosinophils and neutrophils in our dataset because they are found mainly in acute infective or inflammatory conditions, presenting histologically as cryptitis or crypt abscess.

\begin{table*}[t!]
\centering
\begin{tabular}{|l|c|c|c|c|c|c|}
\hline
& \multicolumn{3}{c|}{\textbf{Cross Validation}} & \multicolumn{3}{c|}{\textbf{External Test}} \\
\cline{2-7}  &  \textbf{Binary Dice} & \textbf{Binary PQ} &  \textbf{Multi PQ} &  \textbf{Binary Dice} & \textbf{Binary PQ} &  \textbf{Multi PQ}\\ 
\hline \textbf{U-Net} \cite{ronneberger2015u} & 0.735 $\pm$ 0.028 & 0.515 $\pm$ 0.033 & 0.265 $\pm$ 0.013 & 0.612 $\pm$ 0.087 & 0.390 $\pm$ 0.064 & 0.212 $\pm$ 0.028   \\ 
\hline \textbf{Micro-Net} \cite{raza2019micro} & 0.786 $\pm$ 0.004 & 0.522 $\pm$ 0.015 & 0.264 $\pm$ 0.012 & 0.735 $\pm$ 0.017 & 0.484 $\pm$ 0.019 & 0.244 $\pm$ 0.016  \\ 
\hline \textbf{HoVer-Net} \cite{graham2019hover} & \textbf{0.828 $\pm$ 0.008} & \textbf{0.624 $\pm$ 0.013} & \textbf{0.396 $\pm$ 0.022} & \textbf{0.801 $\pm$ 0.023} & \textbf{0.582 $\pm$ 0.021} & \textbf{0.353 $\pm$ 0.009} \\ 
\hline
\end{tabular}
\caption{Comparative results for each of the baseline models used in our experiments.}
\label{table:results}
\end{table*}

 \begin{figure*}[t!]
		\centering
        \includegraphics[width=1.0\textwidth]{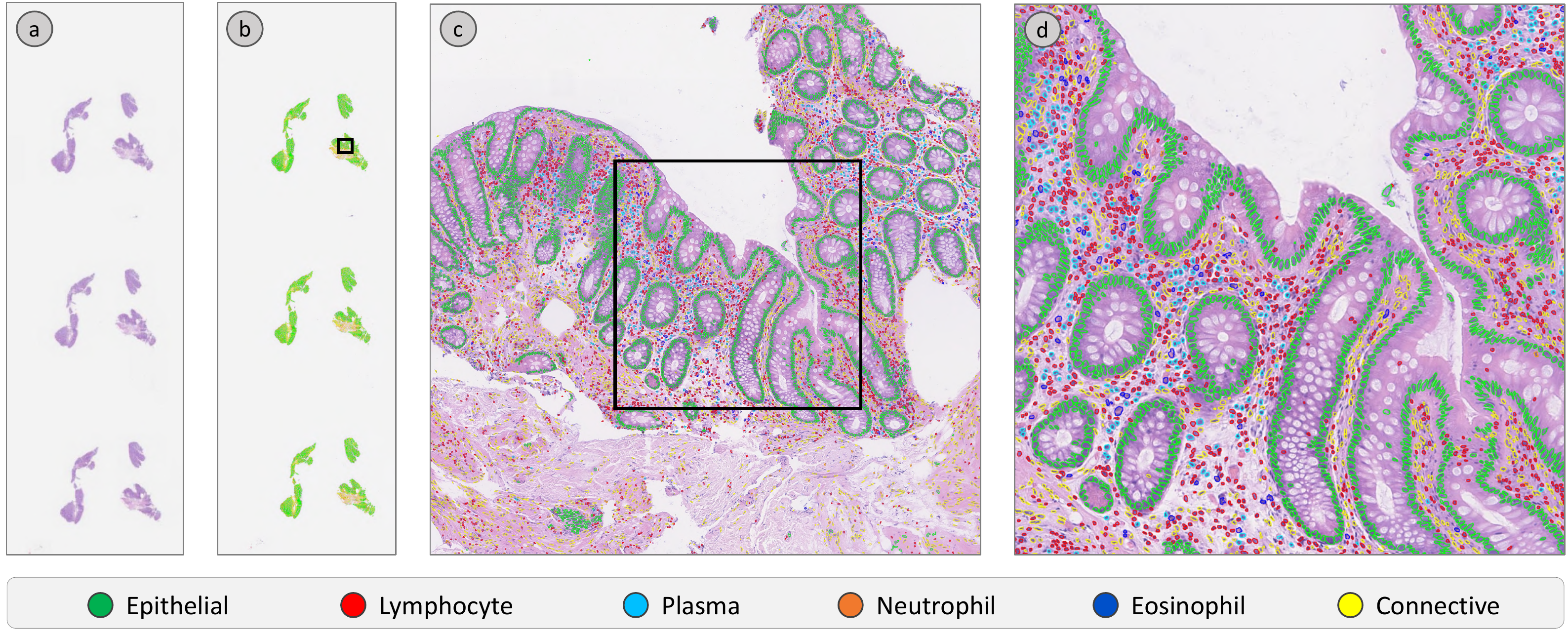}
		\caption{Whole-slide image processed with HoVer-Net. (a) low resolution WSI, (b) low resolution segmentation results, (c) and (d) high resolution segmentation results.} 
		\label{fig:wsi-result}
	\end{figure*}
	
\subsection{Baseline segmentation results}
\subsubsection{Experimental setup}
\label{section:exp-setup}
To enable the comparison of different methods trained on Lizard, we report the results of some baseline experiments. For this, we split the dataset into 3 folds so that cross validation can be performed and the corresponding evaluation statistics computed. The final reported values are the average results across the 3 folds. These splits are provided with the release of Lizard to enable researchers to reproduce our experiments and compare results of new methods. When generating the folds, we ensure that the data is divided in a patient-wise manner. In addition, we use the TCGA part of Lizard as an external test set and report the average of the results processed with each fold-dependent model.
	
In particular, we train U-Net \cite{ronneberger2015u}, Micro-Net \cite{raza2019micro} and HoVer-Net \cite{graham2019hover} on each fold. U-Net is a widely used network for biomedical image segmentation and therefore serves as a strong performance benchmark for this purpose. Micro-Net is a model specifically developed for segmentation tasks in computational pathology, where it was originally applied to the tasks of gland and nuclear segmentation. For both U-Net and Micro-Net we add an extra 1$\times$1 convolution at the output of each network to additionally enable the classification of different nuclear types. HoVer-Net is a state-of-the-art model for simultaneous nuclear segmentation and classification that has been shown to perform well across a large range of datasets \cite{graham2019hover, gamper2020pannuke, verma2021monusac2020}. It is important to note that previous post processing techniques were adapted in line with the resolution of this dataset.

To quantify the performance of each model, we use the following three metrics: $i$) binary Dice score to measure the separation of all nuclei from the background; $ii$) binary panoptic quality and $iii$) multi-class panoptic quality. Originally proposed by \cite{kirillov2019panoptic} and advocated by Graham \textit{et al.} \cite{graham2019hover}, panoptic quality (PQ) for nuclear instance segmentation is defined as:
\begin{equation}
	\small
	\mathcal{PQ}= 
	\underbrace{\frac{|TP|}{|TP|+\frac{1}{2}|FP|+\frac{1}{2}|FN|}}_{\text{Detection Quality(DQ)}}
	\times
	\underbrace{{\frac{\sum_{(x,y)\in{TP}}{IoU(x,y)}}{|TP|}}}_{\text{Segmentation Quality(SQ)}}
	\end{equation}
where \textit{x} denotes the ground truth mask, \textit{y} denotes the segmentation mask and IoU denotes intersection over union. Each (\textit{x,y}) pair is mathematically proven to be \textit{unique} \cite{kirillov2019panoptic} over the entire set of prediction and GT segments if their $IoU(\textit{x,y})>\text{0.5}$. The unique matching splits all available segments into matched pairs (TP), unmatched GT segments (FN) and unmatched prediction segments (FP) and the PQ can be subsequently computed. Binary PQ assumes all nuclei belong to the same class, whereas for multi-class PQ, the score is computed for each class independently and the results averaged to provide us with a unified measure of instance segmentation and classification. 

\subsubsection{Baseline results}
We observe from Table \ref{table:results} that HoVer-Net achives the best performance for both cross validation and on the external test set by a significant margin. Therefore, particularly HoVer-Net serves as a strong performance baseline on this dataset and may potentially be used in downstream cell-based analysis pipelines. Despite using a slightly lower resolution than recent nuclear segmentation datasets \cite{gamper2020pannuke, graham2019hover, verma2020multi}, our results are competitive, which implies that 20$\times$ may be sufficient for nuclear instance and segmentation and classification tasks. As well as reporting quantitative metrics, we display the visual segmentation and classification results of a whole-slide image that has been processed with HoVer-Net and trained on Lizard in Figure \ref{fig:wsi-result}. The sample whole-slide image that we utilise in this figure is from a different source that has not been used during training the model. 

\section{Conclusions and future directions}
In this paper, we introduced the largest known available dataset for nuclear instance segmentation and classification in computational pathology, named Lizard, consisting of nearly half a million annotated nuclei with their associated class labels. Our dataset focusses on colon tissue to ensure that we utilise images from a wide variety of colonic conditions and increase the chance of generalisation to unseen examples. Lizard was generated using a multi-stage pipeline with significant pathologist-in-the-loop refinement steps to enable the collection of accurate annotations at scale. As well as validation of the annotations during dataset generation, we apply a novel sampling method to enable the calculation of concordance measures on a representative sample of the dataset. Furthermore, as part of this work we provide several performance benchmarks to encourage researchers to test their own developed models on our dataset. By making Lizard available, we hope that it will enable the development of accurate and interpretable downstream models for the computational analysis of H\&E stained colon tissue. 

In future work, we wish to expand Lizard and potentially increase the number of considered nuclear types, including macrophages and different epithelial subtypes. In addition, signet ring cells are of particular interest, due to poor prognosis of patients with signet ring cell carcinoma \cite{park2015signet}. Despite epithelial cells displaying varying appearance between different tissues, inflammatory cells will appear similar and therefore Lizard may potentially be further used in non-colonic CPath applications. Next, we wish to utilise the output of HoVer-Net trained on Lizard to develop cell-based models to predict clinical outcome. 
	
\section*{Acknowledgements}
The authors would like to acknowledge the support from the PathLAKE digital pathology consortium which is funded from the Data to Early Diagnosis and Precision Medicine strand of the government’s Industrial Strategy Challenge Fund, managed and delivered by UK Research and Innovation (UKRI).

 \appendix
 \renewcommand\thefigure{\thesection.\arabic{figure}} 
\renewcommand{\thetable}{\thesection.\arabic{table}}

 \section*{Appendix}
 \section{Representative sample cluster examples} 
 \label{section:clusters}
 \setcounter{figure}{0} 
   \begin{figure*}[!t]
		\centering
        \includegraphics[width=1.0\textwidth]{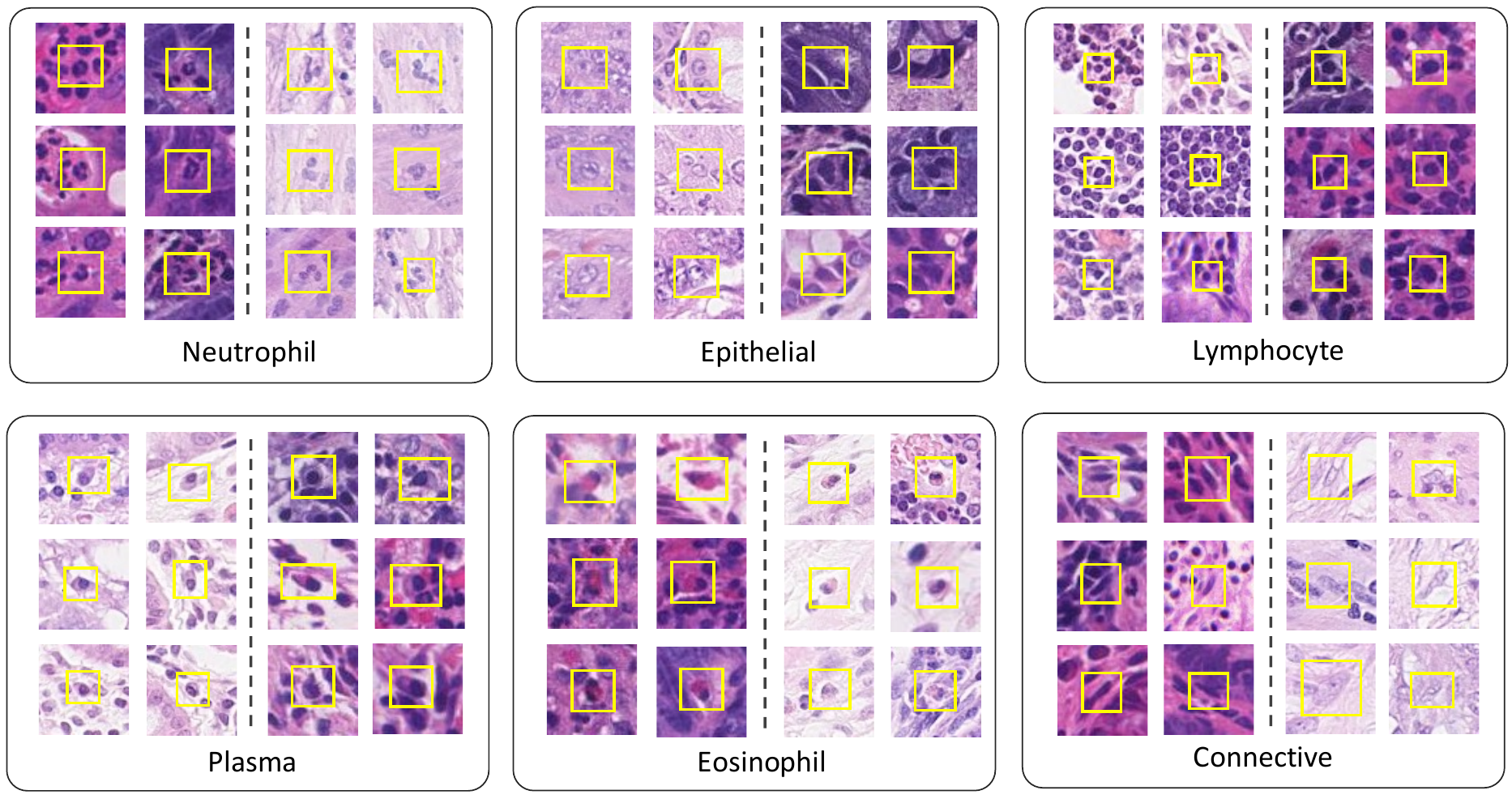}
		\caption{Sample clusters using hierarchical agglomerative clustering. We display 6 example patches per cluster and 2 clusters per category. Within a single category, the different clusters are separated by the vertical dashed line. The yellow boxes highlight each central nucleus.} 
		\label{fig:clusters}
	\end{figure*}
	
 In Figure \ref{fig:clusters} we display clusters obtained using the method as described in Section \ref{section-contrastive}. Specifically, in the figure we show 2 clusters per category, where we display 6 example image patches per cluster. We should expect that for each category, the intra-cluster variability is lower than the inter-cluster variability. This is evident within the figure, where we observe that nuclei within each cluster have a similar appearance. For example the left connective tissue cell cluster has long and thin nuclei with a relatively homogeneous dark stain, whereas the right cluster contains nuclei that appear much paler. Similarly, we observe that epithelial nuclei in the left cluster contain round pale nuclei with visible nucleoli and the right cluster contains darker stained nuclei. Furthermore, we can see that eosinophils in the left cluster have a more prominent eosinophilic cytoplasm as compared to the right cluster. 
 
  \section{Baseline results per category} 
  \label{section:results-breakdown}
  \setcounter{table}{0}
  In Table \ref{table:results} we show the multi-class nuclear segmentation and classification results in terms of average panoptic quality over the classes. However, this does not give us an indication of the per class segmentation result. In Tables \ref{table:results-perclass1} and \ref{table:results-perclass2}, we provide the breakdown of the segmentation result per class to give a better intuition on the performance of each model. In the same way as described in Section \ref{section:exp-setup}, we report the average 3 fold cross validation results in Table \ref{table:results-perclass1} and the average results on the external test set in Table \ref{table:results-perclass2}. 
  
  The performance of epithelial, lymphocyte and connective tissue classes is quite strong, especially for HoVer-Net. This is to be expected, due to the large number of examples that were provided in our dataset. However, we observe that the PQ for the neutrophil and eosinophil classes are significantly lower than the other classes. We hypothesise that this may be due to the large class imbalance, as can be seen from the dataset distribution provided in Figure \ref{fig:dist}. Despite this, the performance of HoVer-Net for these two classes is slightly better than U-Net and Micro-Net because dice loss is used, which can help alleviate problems due to class imbalance. Also, for eosinophils we additionally annotate the eosinophilic cytoplasm which may make it more challenging for the algorithm when most other classes only consider the nucleus. We must also note that for all experiments, we train with a smaller proportion of the entire dataset, due to the usage of 3 folds and an external test set. When training with the entire dataset, the performance of the under-represented classes should improve due to the increase in the number of provided examples. 
 
 \begin{table*}[t!]
\centering
\begin{tabular}{|l|c|c|c|c|c|c|}
\hline
\cline{2-7}  &  \textbf{Epithelial} & \textbf{Lymphocyte} &  \textbf{Plasma} &  \textbf{Neutrophil} & \textbf{Eosinophil} &  \textbf{Connective}\\ 
\hline \textbf{U-Net} \cite{ronneberger2015u} & 0.432 $\pm$ 0.035 & 0.498 $\pm$ 0.036 & 0.175 $\pm$ 0.033 & 0.024 $\pm$ 0.022 & 0.082 $\pm$ 0.045 & 0.377 $\pm$ 0.028   \\ 
\hline \textbf{Micro-Net} \cite{raza2019micro} & 0.437 $\pm$ 0.016 & 0.504 $\pm$ 0.031 & 0.153 $\pm$ 0.045 & 0.017 $\pm$ 0.026 & 0.058 $\pm$ 0.017 & 0.415 $\pm$ 0.014  \\ 
\hline \textbf{HoVer-Net} \cite{graham2019hover} & \textbf{0.559 $\pm$ 0.014} & \textbf{0.577 $\pm$ 0.048} & \textbf{0.365 $\pm$ 0.039} & \textbf{0.158 $\pm$ 0.043} & \textbf{0.219 $\pm$ 0.039} & \textbf{0.499 $\pm$ 0.008} \\ 
\hline
\end{tabular}
\caption{Panoptic quality per class. Displayed results are the average across 3 folds.}
\label{table:results-perclass1}
\end{table*}

 \begin{table*}[t!]
\centering
\begin{tabular}{|l|c|c|c|c|c|c|}
\hline
\cline{2-7}  &  \textbf{Epithelial} & \textbf{Lymphocyte} &  \textbf{Plasma} &  \textbf{Neutrophil} & \textbf{Eosinophil} &  \textbf{Connective}\\ 
\hline \textbf{U-Net} \cite{ronneberger2015u} & 0.266 $\pm$ 0.096 & 0.432 $\pm$ 0.037 & 0.199 $\pm$ 0.030 & 0.006 $\pm$ 0.006 & 0.081 $\pm$ 0.077 & 0.286 $\pm$ 0.051   \\ 
\hline \textbf{Micro-Net} \cite{raza2019micro} & 0.376 $\pm$ 0.035 & 0.454 $\pm$ 0.026 & 0.270 $\pm$ 0.157 & 0.008 $\pm$ 0.013 & 0.063 $\pm$ 0.025 & 0.371 $\pm$ 0.036  \\ 
\hline \textbf{HoVer-Net} \cite{graham2019hover} & \textbf{0.495 $\pm$ 0.025} & \textbf{0.513 $\pm$ 0.007} & \textbf{0.314 $\pm$ 0.012} & \textbf{0.194 $\pm$ 0.023} & \textbf{0.234 $\pm$ 0.055} & \textbf{0.494 $\pm$ 0.029} \\ 
\hline
\end{tabular}
\caption{Panoptic quality per class on the external test set. We process the test set with each fold-dependent model and display the average results.}
\label{table:results-perclass2}
\end{table*}
  \section{Concordance statistics per category} 
  \label{section:concordance-breakdown}
   \setcounter{figure}{0} 
   
\begin{figure*}[t]
		\centering
        \includegraphics[width=0.96\textwidth]{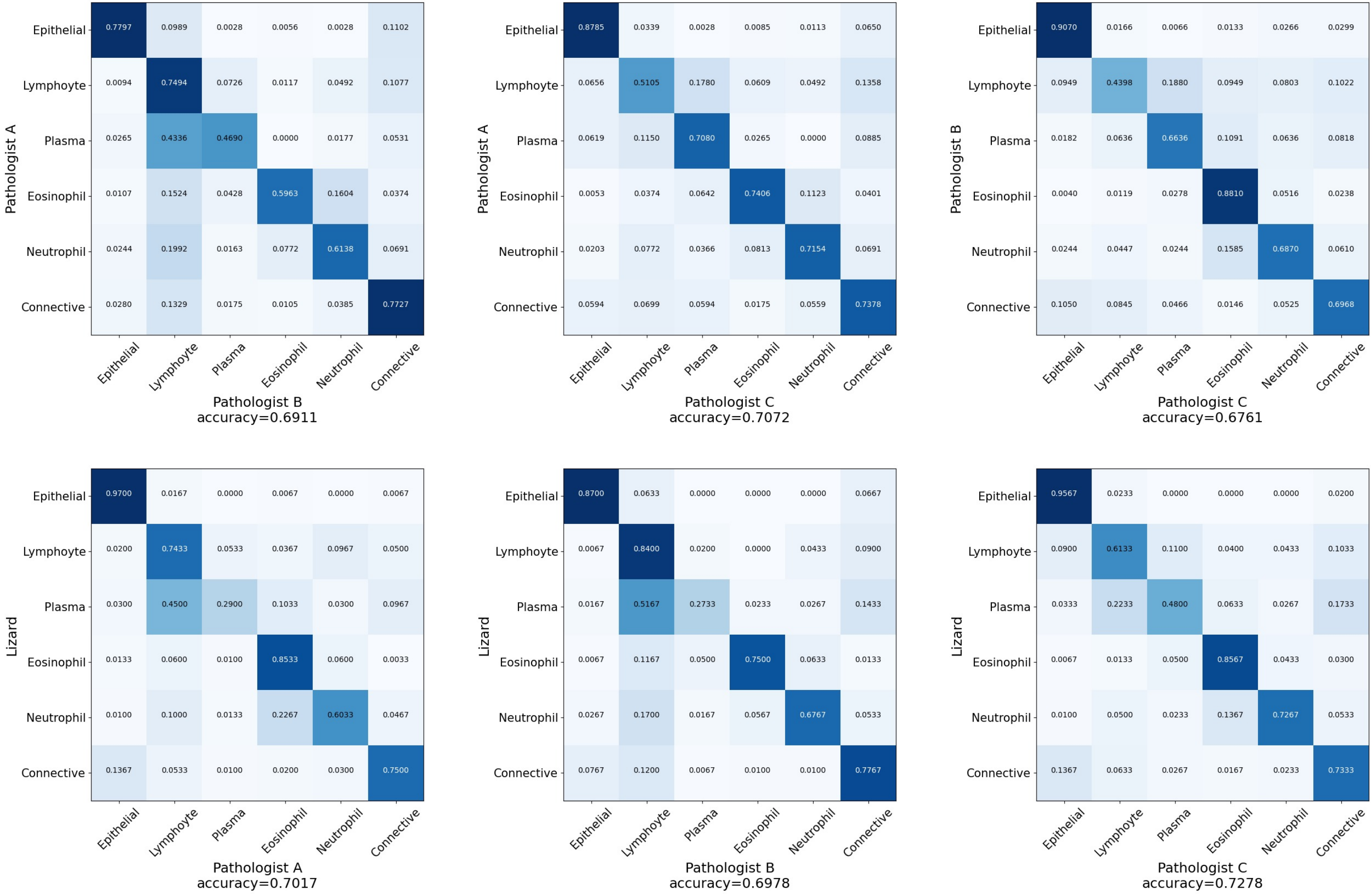}
		\caption{Confusion matrices showing the agreement between two raters for all nuclear categories in our dataset.} 
		\label{fig:conf-concord}
	\end{figure*}

     Despite providing concordance statistics in Table \ref{table:concordance}, these values do not provide us with an idea on the concordance per individual nuclear category. This may be important because it can give us further intuition into where pathologists disagree with each other and with the labels of the dataset. In Figure \ref{fig:conf-concord} we display a confusion matrix for each of the settings shown in Table \ref{table:concordance}, which highlights the degree of agreement between the categories assigned by the two raters. Perfect agreement would result in high values (close to 1) along the main diagonal. 
     
     Overall, we can see that generally we have high values along the main diagonal, indicating strong agreement between the raters. The best agreement between each rater was for epithelial cells, whereas the main source of disagreement was between lymphocytes and plasma cells. This disagreement was more pronounced between each pathologist and our dataset annotations, which implies that there may be some inaccuracies in the distinction between plasma cells and lymphocytes. However, biologically speaking, these are both chronic inflammatory cells and therefore could potentially be grouped into a single category in future work. These confusion matrices also give us an idea of the upper bound of the performance of a classification model trained to distinguish between the 6 categories in our dataset.

{\small
\bibliographystyle{ieee}
\bibliography{egbib}
}

\end{document}